\documentclass{sig-alternate-05-2015}
\usepackage{paralist}
\usepackage{multirow}
\usepackage{array}

\usepackage{soul, color}
\sethlcolor{green}

\begin{document}

%

\conferenceinfo{xx}{xx}
\doi{xx}

\title{An assessment of orthographic similarity measures for several African languages}


\numberofauthors{1} 
%
\author{
%
%
\alignauthor
C. Maria Keet\\
       \affaddr{Department of Computer Science}\\
       \affaddr{University of Cape Town}\\
       \affaddr{South Africa}\\
       \email{mkeet@cs.uct.ac.za}
}
\date{30 July 1999}

\maketitle
\begin{abstract}
Natural Language Interfaces and tools such as spellcheckers and Web search in one's own language are known to be useful in ICT-mediated communication. Most languages in Southern Africa are under-resourced, however. Therefore, it would be very useful if both the generic and the few language-specific NLP tools could be reused or easily adapted across languages. 
This depends on the notion, and extent, of similarity between the languages. We assess this from the angle of orthography and corpora. Twelve versions of the Universal Declaration of Human Rights are examined, showing clusters of languages, and which are thus more or less amenable to cross-language adaptation of NLP tools, which do not match with Guthrie zones. To examine the generalisability of these results, we zoom in on isiZulu both quantitatively and qualitatively with four other corpora and texts in different genres. The results show that the UHDR is a typical text document orthographically. The results also provide insight into usability of typical measures such as lexical diversity and genre, and that the same statistic may mean different things in different documents. While NLTK for Python could be used for basic analyses of text, it, and similar NLP tools, will need considerable customization.
\end{abstract}

{}



\section{Introduction}
\label{sec:intro}

ICTs with an interface in an African language have been on the increase for a multitude of reasons, and multinationals are investing in it. For instance, Google Inc. has their search engine interface in multiple South African languages, such as isiZulu and Setswana, and has a rudimentary Google-translate\footnote{Accessible via \url{https://www.google.co.za/} and \url{https://translate.google.com/}; last accessed: 9-6-2016.} for English-isiZulu since 2013, Facebook\footnote{\url{https://www.facebook.com/translations/}; last accessed: 9-6-2016.}  offers its interface in, among others, Chichewa, Kiswahili, isiZulu, and Shona, and there are localisations of the Ubuntu operating system\footnote{\url{https://translations.launchpad.net/+groups/ubuntu-translators}; last accessed: 9-6-2016.}. This general trend toward more texts in regional languages pushes for natural language processing (NLP) tools to deal (better) with it, such as spellcheckers and a word-completion feature on mobile phones. This increased demand for NLP in under-resourced languages raises the question about the feasibility of cross-language bootstrapping of NLP tools. Isolated experiments have been carried out to that extent. For instance, in the knowledge-driven approach, a morphological analyser developed specifically for isiZulu was used to bootstrap one for Ndebele \cite{Pretorius12} and Setswana \cite{Pretorius09}, a linguistic ontology framework for the noun class system \cite{CK15}, and bootstrapping Runyankore resources from isiZulu \cite{Byamugisha16}. Data-driven (statistical) approaches mainly allude to the hope of transferability across languages \cite{Ndaba16,Spiegler10}, or obtaining only  limited to modest success; e.g., \cite{Baumann14}, in searching for isiZulu affixes, misses prefixes (e.g., {\em ulu-}) and 1/3 of the mere 9 suffixes found were not isiZulu suffixes. 

Underlying these works on transferability and bootstrapping is the assumption  of sufficient linguistic similarity---however determined---across in what is, in linguistics, still called the Bantu language family. The desire to find a general common core across Bantu languages has taken many forms and arguments over the years, and a somewhat more modest version of it is to consider at least `clusters' of languages as one language  with multiple dialects (e.g., isiZulu, isiXhosa, and Ndebele). This has been motivated primarily from a linguistics perspective, such as Meinhof's classification of the noun class system with adjustments tailored for each Bantu language. However, it may also serve cross-fertilisation of computational tools for natural language processing across languages, if considered more broadly. For instance, to speed up the development of  spellcheckers, multilingual search, and machine translation, among many NLP application areas.

This raises multiple questions on cross-language reuse, or at least bootstrapping, of NLP tools as well as a possible data-based approach cf. a knowledge engineering approach. 
We aim to contribute to shedding light on language similarity---hence, potential for reusability of tools across languages---using an approach availing of orthography and representativeness of corpora and texts, which more resourced languages typically rely on to learn tools such as spellcheckers and grammars. We shall answer the following questions:
\begin{compactenum}
	\item Is the orthography across Bantu languages merely a distinction between disjunctive and agglutinating?
	\item Are the orthographic differences, if any, statistically significant?
	\item In using a corpus-based approach, can 1) small corpora be useful as a data source for learning, 2) existing typical NLP measure easily be reused for the Bantu language family?
\end{compactenum}
To answer the first two questions, we compare the text characteristics of a document available in several Bantu languages, namely the Universal Declaration of Human Rights (UDHR). 
The results show that while there are clusters of languages with agglutinating orthography and (highly) disjunctive orthography, there are also languages in-between that have a statistically significant distinct pattern. To validate generalisability of this outcome on a small text document, we zoom into isiZulu to both quantitatively and qualitatively assess corpus and text document characteristics and therewith answer question 3. The UDHR exhibits characteristics typical of isiZulu texts, hence, can be considered representative orthographically. Examining the contents in more detail, one can observe variation in characteristics for different  genres, as for other languages.  Further, the qualitative assessment induced some lessons learnt for some data-oriented approaches and typical corpus statistics measures, 
and we base recommendations on the data analysed. 

The remainder of the paper is structured as follows. We first outline the methodology with materials and methods in Section~\ref{sec:method}. Subsequently we present the results in Section~\ref{sec:main}, which are discussed and compared to related works in Section~\ref{sec:disc}. We conclude in Section~\ref{sec:concl}.

\section{Methodology}
\label{sec:method}

The main aim of the experimental approach is to answer the research questions described in Section~\ref{sec:intro}. The overarching approach to achieve this is to use methods of Small Corpus Studies with its characteristic of `Early Human Intervention' (see \cite{Ghadessy01} for details).

\subsection{Methods}
Two experiments will be conducted. The first experiment compares orthography using one type of document shared across several Bantu languages. The second experiment delves deeper into small corpora and texts for one Bantu language in particular, being isiZulu, which is part of the Nguni language cluster (along with isiXhosa, Ndebele, and siSwati) and first/home language of about 23\% of the population in South Africa.

\subsubsection{Text comparisons across languages} The first step is to select a document available in multiple languages, based on spanning different geographic regions and Guthrie zones \cite{guthrie1971comparative}, national interest, and availability of a document in a language of that zone. The UDHR satisfies these requirements. Data processing includes: computing word length distributions of the words in the documents for each language, with a plain and a cumulative frequency distribution, and other factors, such as the final vowel rule. From the outcome of the exploratory data analysis and descriptive measures, select the appropriate statistical tests to test for significance on differences in orthography, which serves as a measure of relatedness regarding the property of disjunctive-ness/agglutination.

\subsubsection{Corpora and texts in isiZulu} Collect corpora and texts in isiZulu, and clean data where necessary. Compute usual measures such as their size, cumulative relative frequency, and lexical diversity. The lexical diversity is calculated as the ratio of types to tokens. Analyse the texts and corpora on type of words (nouns, verbs, other), their meaning, and any errors and similar confounding factors having to do with Bantu language-specific features, such as the agglutination.

\subsection{Materials: Corpora, text documents, and software}

Only one isiZulu text corpus is freely available, being Ukwabelana (UC), which is composed of an old translation of the bible and a few fiction novels \cite{Spiegler10}; this can be considered of the `fiction' genre. Two readily available text documents were selected, being the UDHR \cite{UDHR} and the Constitution of the Republic of South Africa (SAC), which are manually translated quality texts, and of the genre `public administration/government' texts. The Apache OpenOffice isiZulu spellchecker\footnote{\url{http://extensions.openoffice.org/en/project/zulu-spell-checker}; last accessed: 9-6-2016.}  uses several wordlists, which have been combined into one (OOspell). They contain the wordlist of the bible in isiZulu, medical terms from an isiZulu-English medical dictionary, government text, South African postcodes, and a list of frequent words. Finally, a small news item corpus from \cite{Ndaba16} was used, consisting of news articles from the online versions of {\em Isolezwe} and {\em isizulu.news24} over a time period of August-September 2015 (NIC). An IsiZulu National Corpus is under development \cite{Khumalo15}, but access to the full text corpus was not available, and therefore not included.

	The software to compute the metrics is the NLTK toolkit \cite{nltk09} (importing {\tt string}), which is a set of Python modules for analysis text documents and corpora, in particular its {\tt len} (length), {\tt lexical\_diversity} (or type to token ratio), and {\tt ConditionalFreqDist} for the conditional frequency distribution for the UDHR analysis. It comes with several corpora (as {\tt nltk.data}), including the UDHR in many languages. These manually translated versions of the UDHR have been used for the analysis across languages. For the top-k, vowel-ending words, successive vowels, and `r' presence, a separate Python script was written.

	Statistical analyses are carried out with MS Excel and the more usable online statistical hypothesis tests apps, in particular: Shapiro-Wilk to test for normality of the data set \cite{ShapiroWilk}, Kruskal-Wallis for multiple datasets that are not normally distributed \cite{KruskalWallis}, Mann-Whitney for two non-normally distributed datasets \cite{MannWhitney}, and $\chi^2$ for the test of independence of (multiple) categorical data sets \cite{chi2}.

\begin{figure*}[t]
\centering
   \includegraphics[width=1.0\textwidth]{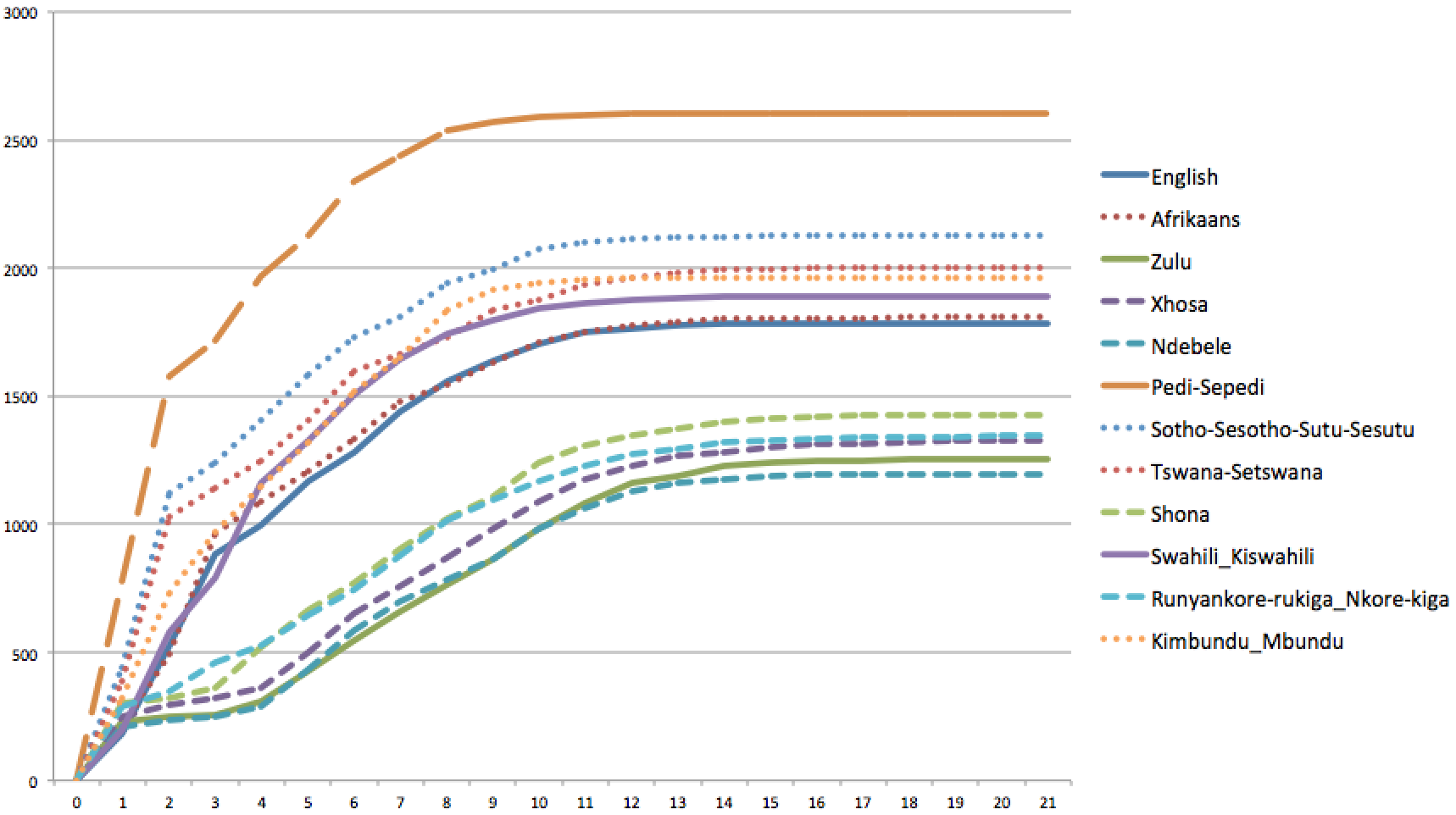}
    \caption{Cumulative frequency distributions of the words in the UDHR of several languages spoken in Sub-Saharan Africa.}
    \label{fig:cfdLang}
\end{figure*}

\section{Results}
\label{sec:main}

We present the results of the UDHR analysis first, then the analysis of the isiZulu texts.

\subsection{Comparing orthography across several  languages spoken in Africa}

The languages selected for analysis of the respective \linebreak UDHRs (named as in {\tt nltk.data}) are listed in Table~\ref{tab:langs}, together with their Guthrie zone classification (where applicable) and basic statistics of the text, and English and Afrikaans for comparison. Regarding Guthrie zones, note that while the Sx languages are in the same zone, there is still some 2000-3000km between the predominant isiXhosa-speaking region (the Cape) and Shona-speaking region (centred in Zimbabwe). Swahili and Runyankore are neighbouring zone languages. The H-zone (Kimbundu) lies in the west of Southern Africa.

\begin{table}[h]
\small
\centering
\caption{Languages used for the UHDR text analysis, and their Guthrie zones.} 
\begin{center}
\begin{tabular}{|p{3cm}|p{2cm}|p{2cm}|}
\hline
{\bf Language} &  {\bf Guthrie Zone} & {\bf Size (tokens)}   \\ \hline   \hline    
English &	N/A &	1781 \\ \hline
Afrikaans &	N/A	& 1807\\ \hline
Zulu	 & S (S42) &	1251\\ \hline
Xhosa &	S (S41)	& 1324\\ \hline
Ndebele &	S (S44) &	1194\\ \hline
Pedi-Sepedi	& S (S33?)	& 2606\\ \hline
Sotho-Sesotho-Sutu-Sesutu &	S (S33?) &	2124\\ \hline
Tswana-Setswana &	S (S31a)	 & 2000\\ \hline
Shona &	S (S10) &	1427\\ \hline
Swahili-Kiswahili &	G (G40) &	1887\\ \hline
Runynakore-rukiga\_Nkore-kiga &	JE10A	& 1345\\ \hline
Kimbundu-Mbundu &	H (H20) &	1959\\ \hline
\end{tabular}
\end{center}
\label{tab:langs}
\end{table}

\subsubsection{Word length distributions}

The cumulative frequency distributions of the of the word lengths in the text in the selected languages are shown in Figure~\ref{fig:cfdLang}.
The largest word length was 21 characters, and the smallest 1, the latter being generally due to strings like {\em Isigaba 1} `Article 1', where the numbers count as tokens as well. Pedi/Sepedi had the most tokens, with 2606 tokens, and Ndebele the fewest with 1194 tokens; thus, for the same information content, a Pedi/Sepedi text has more than twice the number of tokens as a Ndebele text. 
 
As can be seen from Figure~\ref{fig:cfdLang}, there is a clustering regarding word length. To determine whether these visual differences are real ones, we conducted several statistical tests. First, in the bottom group of the figure: is the top-most one, Shona, different from Xhosa, Zulu, and Ndebele? The Shapiro-Wilk test determined the data to be not-normally distributed. Using therefore a Kruskal-Wallis test with the following null and alternative hypothesis:
\begin{compactitem}
	\item[$H_0$:] The samples come from populations with equal means  
	\item[$H_a$:] The samples come from populations with different \linebreak means
\end{compactitem}
and a significance level $\alpha$ = 0.05, then $H_0$ is not rejected (p = 0.08), i.e., there is not enough evidence to state that, orthographically, the agglutination is significantly different among these languages. Doing the same for Zulu, Xhosa, Ndebele and one of the languages in the middle region, Afrikaans, with the same significance level, then $H_0$ is rejected, i.e., Afrikaans is (very) significantly different from the others (p = 0.0002).

Performing the same test, Kruskal-Wallis test, for the languages in the middle region of the graph, being Afrikaans, English, Kiswahili, Sotho, and Tswana, then $H_0$ has to be rejected (p = 0.0008); that is, while they all are in some `middle zone' in the figure, at least one of the languages is statistically significantly different. By successive elimination of Sotho and Setswana that are at the higher regions in the graph---i.e., thus only comparing Afrikaans, English, and Kiswahili---we obtain a p value of 0.076, in that then we cannot reject $H_0$, thus that there is not enough evidence to state that, orthographically, the three languages exhibit a different pattern on word length (as proxy for the disjunctive vs. agglutinative nature of the words in the UDHR). Comparing Sotho and Setswana with a Mann-Whitney test, it is of note that they are different amongst themselves as well (p = 0.0003). Finally, evidently, Pedi/Sepedi is an outlier in disjunctive orthography.

From the tokens in Table~\ref{tab:langs} and CFDs in Figure~\ref{fig:cfdLang}, the respective lexical diversities (type-to-token ratio) behave as expected for agglutinating and disjunctive orthography: in the bottom-cluster, they are around 0.5, in the middle cluster around 0.3, and Pedi/Sepedi 0.23. An illustrative example is `and', which is a word in English that appears 102 times (5.7\% of all tokens), whereas in, e.g., isiZulu, this is a phonologically conditioned {\em na} that is attached to the second noun. For instance, {\em sobulungiswa noxolo} `justice and peace' ({\em na + uxolo = noxolo}), where `peace' appears 3 times in English, but we have {\em noxolo}, {\em wexolo}, and {\em uxolo} in isiZulu, counting as three different words, thereby pushing up the lexical diversity value. Also, unlike English, isiZulu does not use articles, yet the English UDHR has 139 `the' tokens (7.8\% of all tokens).

\subsubsection{Other orthographic features}

\begin{table}[t]
\small
\centering
\caption{Other orthographic peculiarities: percentage of tokens that have a vowel as final character, incidence of consecutive vowels, and the number of {\tt r}'s in the document.} 
\begin{center}
\begin{tabular}{|p{4cm}|p{0.9cm}|p{1.3cm}|p{0.5cm}|}
\hline
{\bf Language} &  {\bf \% FV} & {\bf $\mid$2 vowel$\mid$} & {\bf $\mid${\tt r}$\mid$}   \\ \hline   \hline    
Zulu	 & 99.90 & \hfill 0 &	\hfill 3\\ \hline
Xhosa & 97.19 &\hfill	30	& \hfill 12 \\ \hline
Ndebele & 99.37 & \hfill 4	 & \hfill 3	\\ \hline
Pedi-Sepedi	& 95.58 &  \hfill 346	& \hfill 115 \\ \hline
Sotho-Sesotho-Sutu-Sesutu & 89.77 & \hfill94	 & \hfill 44	\\ \hline
Tswana-Setswana & 92.48 & \hfill112	 & \\ \hline
Shona & 97.78 & \hfill 81& \hfill 409	\\ \hline
Swahili-Kiswahili & 99.76 & \hfill280	 & \hfill 126	\\ \hline
Runynakore-rukiga\_Nkore-kiga & 99.40 &\hfill	271	& \hfill 469 \\ \hline
Kimbundu-Mbundu & 99.77 &\hfill	12 & \hfill 0	\\ \hline 
English & 28.13 &\hfill 316 & \hfill 560 \\ \hline
\end{tabular}
\end{center}
\label{tab:langfeat}
\end{table}

 \begin{figure}[h]
\centering
   \includegraphics[width=0.46\textwidth]{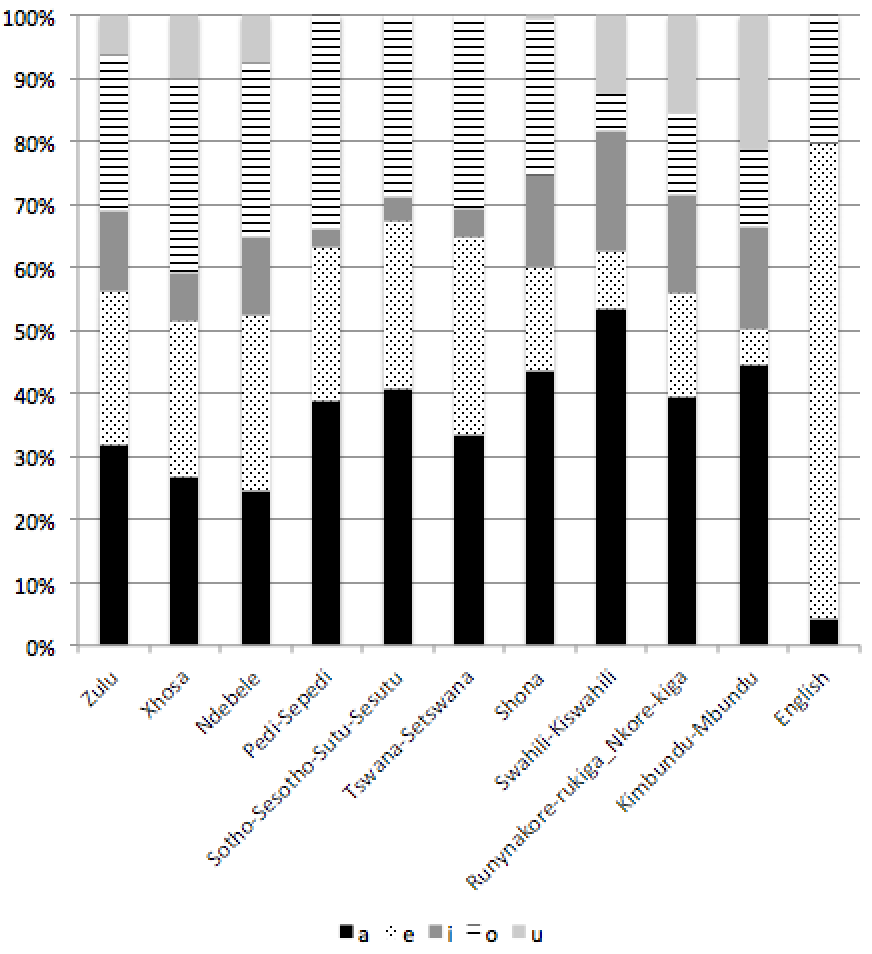}
    \caption{Vowel-ending tokens by language of the words in the UDHR, normalised.}
    \label{fig:vowelDistri}
\end{figure}

Three other typical features we know that generally hold for isiZulu orthography are that isiZulu words (nouns, verbs, adjectives, etc.) have to have a final vowel, words do not have two consecutive vowels, and there is no `r' in the alphabet\footnote{Those words that do have an `r' are loanwords that are not fully assimilated, such as {\em i-okhestra} `orchestra', cf., e.g., {\em ikhompuyutha} `computer'.}. The data obtained is shown in Table~\ref{tab:langfeat}, with a breakdown of the final vowels shown in Figure~\ref{fig:vowelDistri}. As visually it looks like there are `clusters' of languages regarding the final vowel, we subject the data to statistical tests ($\chi^2$). First, $H_0$ and $H_a$ are as follows, for the two categorical variables under consideration:
\begin{compactitem}
	\item[$H_0$:] Vowel-ending distribution for set of languages does not differ significantly.
	\item[$H_a$:] Vowel-ending distribution for a set of languages do differ significantly.
\end{compactitem}
We commence our tests with a $\chi^2$ comprising the  bottom-cluster of Figure~\ref{fig:cfdLang}---isiZulu, isiXhosa, isiNdebele, Shona, and Runyankore. It has a $\chi^2 = 40$, so with the degrees of freedom (df) of 16, we obtain p $<0.001$, i.e., they are statistically significantly different. Visually, especially Shona looks like an outlier and Runyankore somewhat similar, but these remaining four still results in significance with an $\alpha$ of 0.05 ($\chi^2 = 23.41$, df = 12, p = 0.024). IsiZulu, isiXhosa, and isiNdebele do {\em not} differ significantly ($\chi^2=4.6$, df = 8, p = 0.7993). Based on these results and the descriptive statistic in Figure~\ref{fig:vowelDistri}, one can expect the rest: Setswana is different from isiZulu and isiXhosa ($\chi^2 = 16.9$, df = 8, p = 0.0308), whereas Sepedi, Sesotho, and  Setswana are not statistically significantly different from each other ($\chi^2=2.49$, df = 8, p = 0.9622).

Noteworthy is that isiZulu indeed does not have two successive vowels in any word, as expected. IsiNdebele has four, of which one is an untranslated `preamble' and one that looks like an error, {\em ukuthiukholo}, where a space is missing after {\em ukuthi}. IsiXhosa, on the other hand, does have successive vowels with some of its prefixes, resulting in {\em ee} or {\em ii}, such as {\em iimfanelo} `duty' in noun class 10 that has as prefix {\em ii-}. They are 
relatively remarkably similar in this orthographic feature, yet Runyankore is not. Using \cite{Byamugisha16}'s list of noun prefixes and noting tokes such as {\em emiteekatekyere}, it suggests that the stems themselves may have successive vowels, i.e., the core vocabulary permits it.

\subsection{Characteristics and quality of isiZulu corpora and documents}
\label{sec:zulu}

\begin{figure}[h]
\centering
   \includegraphics[width=0.48\textwidth]{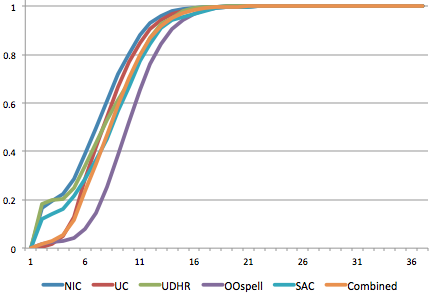}
    \caption{Cumulative relative frequency distribution of the length of the tokens in the isiZulu documents.}
    \label{fig:cfdZu}
\end{figure}

Basic descriptive measures of the selected corpora and text documents are summarised in Table~\ref{tab:langs2}, whereas Figure~\ref{fig:cfdZu} shows the cumulative relative frequency distribution for those corpora and texts. The NIC and the words from OOSpell seem to be furthest apart, yet a Mann-Whitney test with a significance level of 0.05 showed that these differences are not significant at all (p = 0.98). As the UDHR lies somewhere in the middle, it can be concluded that the results obtained with it in the previous section are fairly typical data for text in isiZulu, despite being of a small size.

Also here all documents have tokens of size 1, yet the UC has tokens up to 36 characters. The 1 and 2-character tokens are partially errors and roman numerals, such as {\em kc}, {\em t}, {\em td}, {\em xi}, and {\em zz}, and to some extent also the 3-character tokens (e.g., {\em jsb} and {\em jwi} in the UC sentences, whereas the errors {\em kod}, {\em kae}, and {\em upl} only appear in the word list). The 36-character word is an artefact of the data-centric approach in constructing the corpus, having tokens such as {\em ukungikhombisinqamteyangqubuzumhlaba}, and the individual word list---but not the untagged sentences---even has {\em wathiesholamazwiwabecyisongaincwadicyibeka}, which is clearly a concatenation of different words. This can be seen from unusual successive vowels and a decomposition of its constituents: {\em -isonga-} `save' and {\em -incwadi-} `book', {\em wathi}, {\em esho}, and {\em amazwi} all have to do with `say' and `voice', and {\em ibeka} `put', thus having four verbs in it. Notwithstanding, valid long words exhibiting the strong agglutinative character of isiZulu do exist in the corpora, such as {\em bebengakangikhumbuli} `they had not yet remembered me'. The document with the next-longest token is OOspell's {\em kwakungokokwahlukaniselwa} (25 characters) from its bible-based wordlist and {\em ngokungangesinxephezelo} (23 characters) from the government wordlist. These and other uncommon tokens, notably many words without a final vowel (e.g., {\em sowehlul}[-a] `will beat it', {\em lungikhumbuz}[-a] `it reminds me' in UC), prove that the OOspell and the UC bible text are different versions, with the latter  written in an isiZulu that is, at least, out-dated. 

\begin{table}[h]
\small
\centering
\caption{Basic statistics of the considered corpora and text documents in isiZulu.} 
\begin{center}
\begin{tabular}{|p{3.6cm}|p{1.3cm}|p{2.2cm}|}
\hline
{\bf Corpus/text} &  {\bf Size \newline (tokens)} & {\bf Lexical diversity (rounded)}   \\ \hline   \hline    
News Item Corpus (NIC)	& \hfill 22498 & 0.45 \\ \hline
Ukwabelana (UC) & \hfill 	288106	& 0.30\\ \hline
UDHR & \hfill	1251 &	0.56\\ \hline
OpenOffice Spellchecker files (OOspell)	& \hfill 106450 &	0.84\\ \hline
SA Constitution (SAC) & \hfill 	33056	& 0.24\\ \hline
{\em Combined} &	\hfill{\em 451232}	& {\em 0.36}\\ \hline
\end{tabular}
\end{center}
\label{tab:langs2}
\end{table}

\vspace{-2mm}

\subsubsection{Final vowels}

\begin{figure}[h]
\centering
   \includegraphics[width=0.47\textwidth]{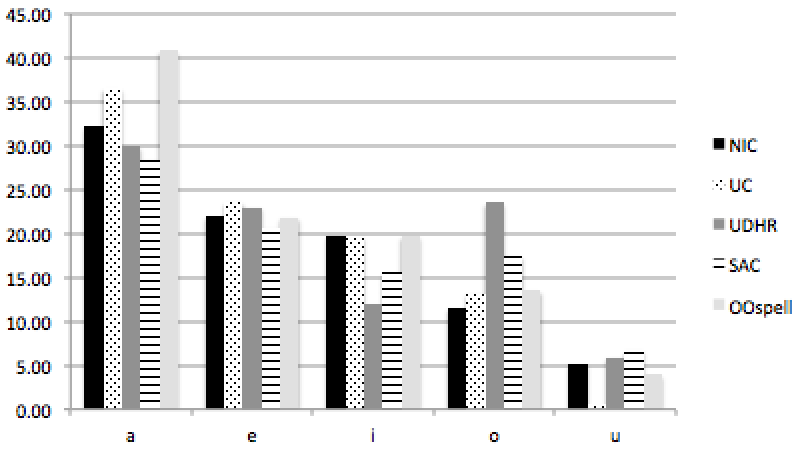}
    \caption{Vowel-ending tokens as a percentage of the total number of tokens in that corpus.}
    \label{fig:vowels}
\end{figure}

The final vowel issue deserved closer inspection, for isiZulu words typically end with a vowel. The basic vowel-ending aggregates are shown in Figure~\ref{fig:vowels}. The single consonant-ending word in the isiZulu UDHR is a loanword, {\em kuCharter}. Besides the general distribution across vowels, such as 7 times as many ending with an {\em -a} than an {\em -u}, it is worth noting the relative outliers with more {\em a}-endings in OOspell, {\em o}-endings in the UDHR, and near-absence of {\em u}-ending tokens in the UC.
From this vowel analysis one can find the deviating consonant-ending words, which are summarised in Table~\ref{tab:fv}. The curated and quality texts of OOspell and UDHR have an extremely low percentage of consonant-ending tokens, such that one relatively safely could design grammars as intended and obtain good performance. The UC and SAC less so, but for different reasons: the SAC has multiple foreign words, whereas the UC has many errors where the words should have a vowel, as noted above with {\em sowehlul}. Finally, there are relatively many consonant-ending tokens in the NIC, because there are many (valid) named entities.

\vspace{-2mm}

\begin{table}[h]
\small
\centering
\caption{Final vowel characteristics with examples; \% c. = percent consonant-ending tokens.} 
\begin{center}
\begin{tabular}{|p{1cm}|p{0.7cm}|p{5.5cm}|}
\hline
{\bf Corpus} &  {\bf \% c.} & {\bf Examples}   \\ \hline   \hline    
NIC	& 9.21 &	{\em uMnuz} `Mr.', {\em iFacebook neTwitter}, {\em EFF}   \\ \hline
UC & 	1.75 	& {\em sowehlul}, {\em ngaphans}, {\em training} \\ \hline
UDHR &	5.73 & 0.1 without article numbers \\ \hline
OOspell	& 0.16 &	{\em Umhlanga Rocks}, {\em uJohannes}\\ \hline
SAC & 	9.57 	& {\em bless}, {\em ANC}, {\em EGauteng} (1.82 without article numbers) \\ \hline
\end{tabular}
\end{center}
\label{tab:fv}
\end{table}

\subsubsection{Lexical diversity}

The lexical diversity is very high for the OOspell file, which is largely due to having combined different word lists, rather than texts, and as such should be disregarded for comparison. Overall, the lexical diversity is high, compared to, say, English. For instance, the lexical diversity of various genres in the Brown corpus is typically in the range of 0.12-0.23 \cite{nltk09}. An important reason for this large difference is due to the simplicity of the measure. For instance, {\em amaphoyisa} (`police', plural) and {\em namaphoyisa} (`and the police', plural) in the NIC are counted as two different words, but semantically refer to one concept; {\em amaphoyisa} occurs only 27 times, yet its root {\em -phoyis-} occurs 47 times in just the first news item set (2657 tokens) of the NIC. There are many more such cases in the corpora, such as {\em -mali} `money', also in the NIC: {\em imali}, {\em kwemali}, {\em yimali}, {\em onemali}, {\em osozimali}, {\em kwezimali}, {\em ngezimali}, which are, respectively of -, and -, that/which/who has -, of - (pl.), about/by/with/per - (pl.) money. A specific instance is included in Table~\ref{tab:fund} for illustrative purpose. In the simple type-to-token measure, they are all counted as different types. This does not explain the substantial difference in lexical diversity between the UDHR and SAC, however, which are of the same genre. The UDHR has a high lexical diversity due to it being a small document. For instance, the subset of the NIC with {\em Isoleszwe} news articles of only August 7, 2015, has a similar lexical diversity of 0.55 on its 2667 tokens. So, this is not unusual.

To get more insight in the possible reasons for the differences in lexical diversity, we now look at the top-20 words in each corpus, their frequency within the corpus, and categorise the top-20 into noun, verb, and other. The results are included in Table~\ref{tab:topk}. It is known that with larger corpora, `auxiliary' words and connectives become more frequent than others, which can be seen from UC's top-20, such as {\em nje} `such/like this', {\em ke} `now, and so, then, very well', {\em ngoba} `because, since'. While in the UC, such words take up 80\% of the top words (16 out of 20), in the UDHR this is only 40\%, and the other two are in-between, as are their sizes. Further, the NIC is about 2/3 the size of the SAC, yet with notable difference in lexical diversity. These are two different genres, and the former has relatively more verbs than nouns, compared to the SAC, which is also the case with that subset of Aug 7. That is, news articles report more on people saying things (a.o., {\em ukuthi}, {\em uthe}, {\em kusho}) than happens in texts stating people's rights (SAC and UDHR). Likewise, stories (UC) also tend to have a centrality on humans ({\em umuntu}, {\em abantu}) saying things ({\em ukuthi}, {\em wathi}). As such, while raw data with the usual numerical analysis may suggest different patterns, only a qualitative analysis of the `early human intervention' approach shows that, from an informational point of view, the document are as in other languages.
\begin{table}[t]
\small
\centering
\caption{Tokens with the {\em -fund-} root in the {\em Isolezwe} articles of August 26 and 27, 2015 (part of the NIC).} 
\begin{center}
\begin{tabular}{|p{2.1cm}|p{0.3cm}|p{4.5cm}|}
\hline
{\bf Token} &  {\bf $n$} & {\bf Translation}   \\ \hline   \hline    
abafundi & 20 & students \\ \hline
bafundi & 6 & students  (note: preceded with {\em laba}, so vowel dropped)\\ \hline
umfundi & 6 & student \\ \hline
nabafundi & 5  & and the students\\ \hline
wezeMfundo & 5 & of those of (an/the) education (note: part of the `Department of Education' phrase) \\ \hline
azifundi & 3 & they do not learn\\ \hline
kubafundi & 3 & in/at/on/to/from (the) students \\ \hline
esifundazweni & 2 & in/at/on/to/from a/the province \\ \hline
kwabafundi & 2 & of (the) students \\ \hline
abafundela& 1& that/which/whom they study/ied for \\ \hline
abafundisa& 1 & teach (note: 3rd. pers. pl.) \\ \hline
abazifundisayo & 1 & verb, several options to decompose \\ \hline
bayazifundisa & 1 & they teach them(selves) \\ \hline
besafunda& 1 & learnt (note: past tense) \\ \hline
efunda &1 & that/which/who learn \\ \hline
ezifundela & 1 & that/which/whom they study for (themselves)\\ \hline
kunomfundi& 1 & it is with (a/the) student \\ \hline
mfundi&1 & student  (preceded with {\em lo}, so vowel dropped\\ \hline
nemfundo & 1 & and knowledge/learning \\ \hline
okunguMfundisi& 1 & that/which is (a/the) teacher (note: as title of a person) \\ \hline
sifundisa&1 & teach (note: 1st pers. pl.)\\ \hline
umfundisi& 1 & teacher \\ \hline
uMfundisi&1 & teacher (note: as title of a person) \\ \hline
wabafundi & 1 & of (the) students (note: PC wa-)\\ \hline
yabafundi& 1& of (the) students (note: PC ya-)\\ \hline
zingafundi & 1 & they are not learning\\ \hline

\end{tabular}
\end{center}
\label{tab:fund}
\end{table}	

\begin{table*}[t]
\small
\centering
\caption{Top-20 words in the corpora and text documents, with their frequency (\#), and percentage (pct.) of the total amount of tokens. Green: (conjugated) verb; yellow: noun; red: either; no colour: auxiliary word.} 
\begin{center}
\begin{tabular}{p{16.6cm}}
\centering
   \includegraphics[width=0.95\textwidth]{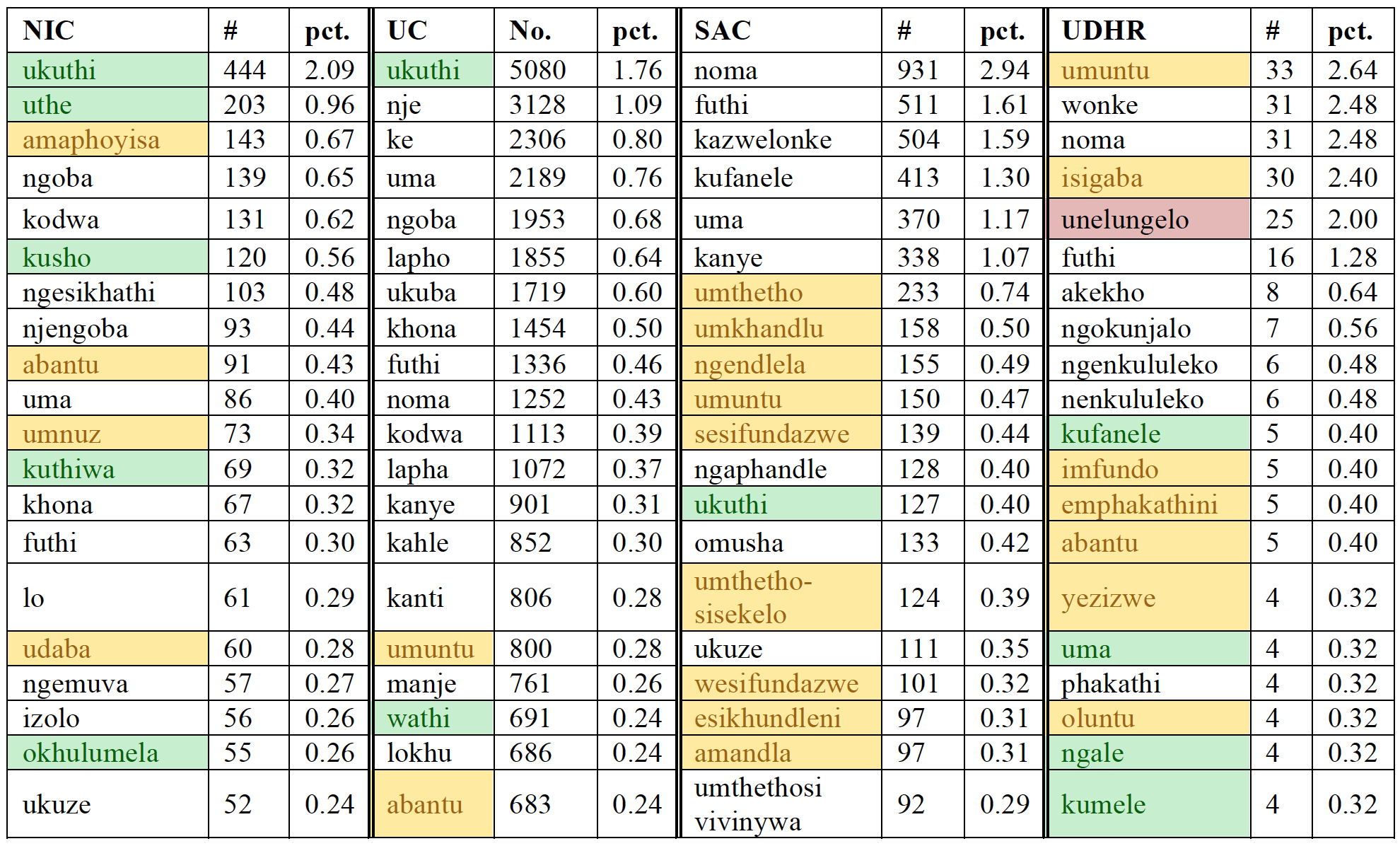}
\end{tabular}
\end{center}
\label{tab:topk}
\end{table*}


\section{Discussion}
\label{sec:disc}

We first return to the research questions as described in Section~\ref{sec:intro} and subsequently discuss several aspects of the data-driven approach with its measures and tools.

\subsection{Answering the research questions}

The first two questions, {\em is the orthography across Bantu languages merely a distinction between disjunctive and agglutinating?} and {\em are the orthographic differences, if any, statistically significant?} has to be answered in the negative for the former and affirmative for the latter. The `negative' is interesting, however, for it revealed that there are at least several languages  somewhere `in-between' of being highly disjunctive or agglutinative. For those languages that are `in-between', it is not just a case of writing the prefixes to a stem disjunctively or together, but only some of the parts of speech or concords are, and then it is likely that for different languages, different choices have been made as to what to write separately and what together. The consequence of this is that, despite a promising case study \cite{Pretorius09}, it is not at all clear whether that particular bootstrapping approach is reusable for other Bantu languages and that instead new rules have to be devised each time. On the positive side, that languages in the lower cluster in Figure~\ref{fig:cfdLang} are not statistically significantly different does indicate good prospects of reusability of techniques with comparatively little adaptation not just between the three know to be similar languages---isiZulu, isiXhosa, and Ndebele---but also Shona and Runyankore. This holds as well for the presence of final vowels of a word, though less so the distribution among the vowels. Bootstrapping prospects will be much less so for Swahili language resources for, say, isiZulu tool development. Further, the assumption that languages in the same Guthrie zone behave the same orthographically cannot be assumed. Therefore, in aiming to reuse resources, it is prudent to first examine whether a rough notion of similarity does exist. The measures used here may assist in that.

The third question considered the possible generalisability of the results that were obtained with a small text document, zooming in on one language, isiZulu: {\em In using a corpus-based approach, can 1) small corpora be useful as a data source for learning, 2) existing typical NLP measures easily be reused for the Bantu language family?} The UDHR itself was in line regarding basic document statistics in relation to the other isiZulu texts considered, showing that small corpora can be useful as a data source for learning. The additional analysis of the other corpora further contributes to supporting the validity of the answer to the first two questions. Further, the UDHR can be considered a `cleaner', high quality document that will result in better accuracy for rule-based approaches to NLP compared to the UC and NIC. There are some limits to the usability of existing typical NLP measures, notably lexical diversity (as type-to-token ratio), and it has been shown that other orthographic aspects provide additional insights, such as the rule on the final character of a word.

\subsection{Issues with measures and data}

While several measures were used successfully, such as the cumulative frequency distribution, number of tokens, and language peculiarities such as the final vowel, the notion of lexical diversity was rather problematic and the non-UDHR texts had some limitations, which are discussed in the remainder of this section. 

\subsubsection{Lexical diversity}

The usual notion of lexical diversity (including permutations \cite{DeBoer14}) and, similarly, word frequency profiling with the log likelihood \cite{Rayson00}, are not informative as measures for agglutinating languages. The {\em amaphoyisa} and {\em imali} were but two examples to illustrate the issue with nouns, where notably prepositions are merged with the noun, and {\em bebengakangikhumbuli} as illustrative for verbs, which contains also concords for subject, object, and others, such as aspect. While this may seem obvious to a linguist, to the best of our knowledge, no agglutinating language-specific lexical diversity formulae for agglutinative languages have been proposed and tested computationally. A possibly relevant, and tried, approach is to use morphological analysers to extract the stem or root \cite{Spiegler10,Pretorius03fsm}. Just categorising by root is not a viable alternative either, however; e.g., {\em -fund-} is the root of {\em abafundi} `students', of {\em umfundisi} `teacher', and of {\em bayazifundisa} `they teach them(selves)', i.e., the same root becomes slightly  different concepts or part of speech depending on the affixes, hence, that would result in over-generalisations in both the lexical diversity and log likelihood values. It may be useful to devise formulae or cookbook-level `preprocessing' steps that are tailored to agglutinating languages so as to obtain meaningful data not only in qualitative assessments, 
but also, moreover, in larger corpora so as to compute a sort of a {\em semantic lexical diversity} or an {\em agglutination-calibrated lexical diversity} (cf. other variants \cite{DeBoer14,Rayson00}). Although we cannot possibly determine this here, it serves to explore this option for further investigation, for a possible chance to reuse the wealth of existing NLP tools. Let us take  `calibration' as example. One can figure out a ratio of `base word' (e.g. {\em abafundi}) to `modified word' (e.g., {\em nabafundi}) with the same meaning but with auxiliaries agglutinated 
 for a text, a genre, or in a language model as a feature of the language\footnote{Assume one can extract the nouns with a 100\% accuracy. Although  this is not possible at the time of writing, POS tagging is being looked into (e.g., \cite{Spiegler10,Pretorius03fsm}).}. Then, with corpus $C$, tokens $T$, set $s$ and the simple type-to-token ratio for a corpus,
\begin{equation}
TT_C = \frac{\mid s(T)\mid}{\mid T\mid}
\end{equation}
it would be modified as follows.  Let us have a language model where $B$ is a base word, $M$ the modified tokens that generally appear in some typical text of language $L$,  and $\beta$ and $\mu$ for their types, taking the median to cater for the long tail distribution:
\begin{equation}
\lambda_t = \mbox{{\sc med}}( \frac{B}{M})
\end{equation}
\begin{equation}
\lambda_{\theta} = \mbox{{\sc med}}( \frac{\beta}{\mu})
\end{equation}
Then the calibrated type-to-token ratio for a corpus would be:
\begin{equation}
TT_{cal} = \frac{\lambda_{\theta} \mid s(T)\mid}{(1 - \frac{1}{\lambda_t})  \mid T\mid}
\end{equation}
 Obviously, one can also calibrate in the opposite direction, from disjunctive to agglutinative. 
  
 To illustrate this with an actual example, let us take tokens with {\em -fund-} in the {\em Isolezwe} articles of August 26 and 27, 2015, which are listed in Table~\ref{tab:fund}: {\em abafundi} `students' is the base noun, and {\em bafundi}, {\em nabafundi}, {\em kubafundi}, {\em kwabafundi}, {\em wabafundi}, and {\em yabafundi} are the modified nouns, thus standing in a ratio of 1:6 as types and 10:9 as tokens, and in the singular as {\em umfundi} with {\em kunomfundi} and {\em mfundi} as 1:2 and 3:1, respectively. This could be done likewise for all words, and taking the median over it to obtain  $\lambda_{\theta}$ and $\lambda_t$, respectively. If there were only these two, then $\lambda_{\theta}$ would be 0.50 and $\lambda_t$ 3.
Filling this into the equation results in a calibrated lexical diversity of 
\begin{equation}
TT_{cal} = \frac{0.50 * 2105}{0.67 * 3774} = 0.42
\end{equation}
compared to the original $\frac{2105}{3774} = 0.56$.

To figure this out systematically for text documents, individual corpora, by genre, or even of the langauge, much research is yet to be carried out for all languages in the Bantu language family.

\subsubsection{Provenance of the text}
An issue with the data-driven approach is the `dirty data' that skews the results, such as the word length, as mentioned in Section~\ref{sec:zulu}, which is beyond the scope of this paper. Its effects have to be considered in the evaluation of corpus-based NLP tools, however, and it was clear that a high quality text such as the UDHR exhibit typical language characteristics, such as final vowels of a word, very low incidence of `r' and, and absence of successive vowels in a word. 

While a measure of `cleanliness' is whether the tokens adhere some basic orthographic rules, such as words ending in a vowel, this should be used with caution. It may simply be an artefact of the source text---genre or datedness---that is included in the corpus rather than `dirtiness': just because the NIC had a much higher percentage of consonant-ending words, this does not imply it is `dirty', but instead had many non-assimilated named entities. Whether this has an effect on corpus-based NLP tools, such as a spellchecker \cite{Prinsloo04} or morphological analyser, remains to be seen in practice. In theory, it certainly does: an automaton or context-free grammar that only accepts strings whose final character is a vowel will do worse on the NIC than on OOspell, UC, or the UDHR. Likewise, a named entity recogniser may be more beneficial for a corpus in the news genre than for others.

\subsubsection{Tooling}

While indeed a general-purpose package, such as the \linebreak NLTK, could be used to obtain basic analyses at least, there are limitations. Its regular grammar feature is woefully inadequate for the complex morphological rules, for instance, because it requires the components already to be split, which is precisely one of the computational challenges yet to be resolved. More generally, there are limitations to reusability of NLP tools and measures that require substantial customisation to handle specifics of Bantu languages, such as a way to compute the real log likelihood or how to adjust the lexical diversity calculations, of which the `calibrated' one was but one possible example. 

It also demonstrates the need for a more generic, larger, corpus as well as one separate-able by genre, and annotations as to the provenance of its source text. These requirements and recommendations reformulate Sharma Grover et al's outcome of the human languages technology audit \cite{Sharma11}, in that there still is a large gap to fill on information and knowledge processing, even 5 years since the audit.

\section{Conclusion}
\label{sec:concl}

The comparison of a shared-information-content document, the Universal Declaration of Human Rights, demonstrated that regarding orthography, there are at least three statistically significant different groups of Bantu languages, which do not match Guthrie zone. It showed potential for easy bootstrapping among several of the languages tested (isiZulu, isiXhosa, Shona, Runyankore), but not others (Swahili, Kimbundu). The UDHR itself is, while a small text, typical for a text in that language, as demonstrated for isiZulu. Further analyses of corpora and text documents showed that: 1) lexical diversity is not a useful measure for agglutinating languages, 2) corpora may need to be cleaned manually, 3) normal grammar rules, such as that a word should end with a vowel, can have a considerable number of valid exceptions, and 4) genre differences were detected that would be good to take into account in future corpus-based NLP tools.

\subsubsection*{Acknowledgements} This work is based on the research supported in part by the National Research Foundation of South Africa Grant Number 93397.

\bibliographystyle{abbrv}
\bibliography{scexperiments}  
%
%
\end{document}